\documentclass[conference]{IEEEtran}
\usepackage[utf8]{inputenc}
\IEEEoverridecommandlockouts
\usepackage{cite}
\usepackage{amsmath,amssymb,amsfonts}
\usepackage[colorinlistoftodos]{todonotes}
\usepackage{bm}
\usepackage{listings}
\usepackage{graphicx}
\usepackage{float}
\usepackage{textcomp}
\usepackage{xcolor}
\usepackage{booktabs}
\usepackage{caption}
\usepackage{subcaption}
\def\BibTeX{{\rm B\kern-.05em{\sc i\kern-.025em b}\kern-.08em
    T\kern-.1667em\lower.7ex\hbox{E}\kern-.125emX}}
    
\newfloat{lstfloat}{htbp}{lop}
\floatname{lstfloat}{Algorithm}
\makeatletter
\let\@oldmaketitle\@maketitle% Store \@maketitle
\renewcommand{\@maketitle}{\@oldmaketitle% Update \@maketitle to insert...
\setcounter{figure}{0}
\begin{center}
    \includegraphics[width=2\columnwidth]{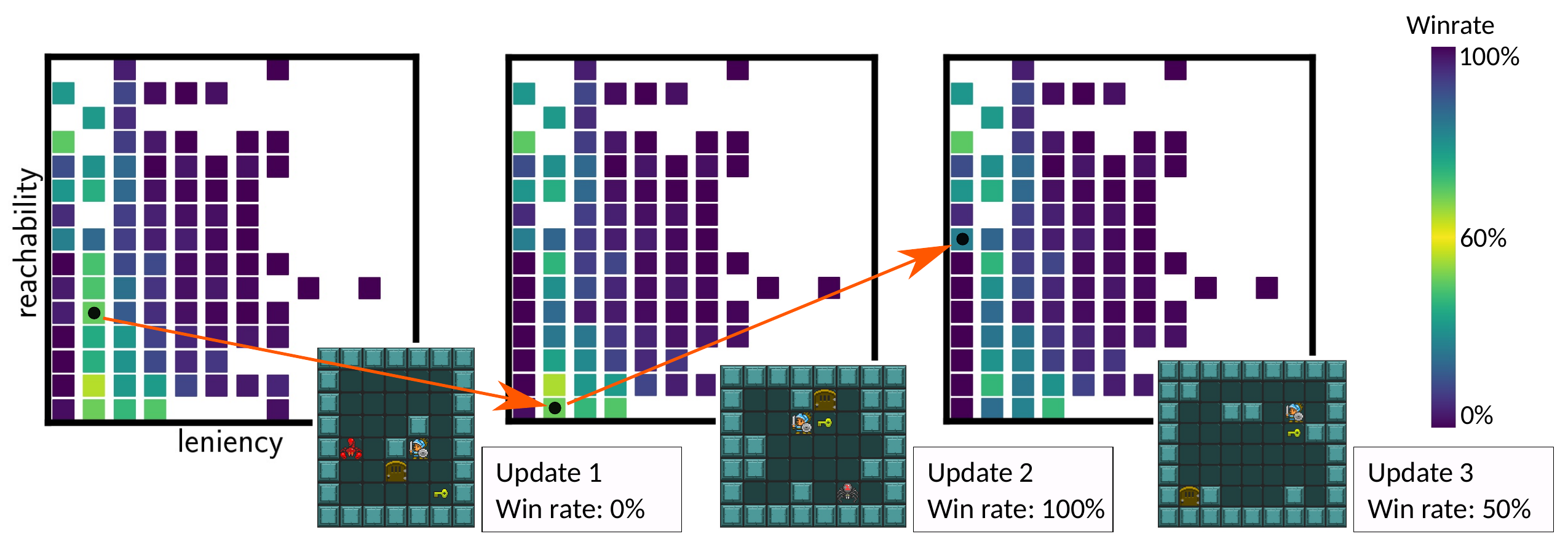} %sweet spot for 
    \captionof{figure}{\textbf{Finding a level with the right difficulty via Intelligent Trial-and-Error (IT\&E).} IT\&E \cite{cully2015robots} for games first creates a set of levels, 
     arranged in a map that varies across level characteristics (amount of enemies and distance to goal). 
    IT\&E updates   its beliefs about the difficulty of each level continuously using Gaussian Processes. In only three updates, the IT\&E algorithm finds a level with ideal difficulty (win rate between 50\%--70\%) for a \textit{One Step Look-Ahead} agent. The brighter the color in the maps, the closer the level is to the target difficulty, with darker colors representing levels that are either too easy or too hard. 
    }\label{fig:updates}
\end{center}
}
\makeatother

\begin{document}

\title{Finding Game Levels with the Right Difficulty in a Few Trials through Intelligent Trial-and-Error}

\author{\IEEEauthorblockN{Miguel González-Duque}
%\IEEEauthorblockA{\textit{Digital Design} \\
\textit{IT University of Copenhagen}\\
Copenhagen, Denmark \\
migd@itu.dk
\and
\IEEEauthorblockN{Rasmus Berg Palm}
\textit{IT University of Copenhagen}\\
Copenhagen, Denmark \\
rasmb@itu.dk
\and
\IEEEauthorblockN{David Ha}
\textit{Google Brain}\\
Tokyo, Japan \\
hadavid@google.com
\and
\IEEEauthorblockN{Sebastian Risi}
%\IEEEauthorblockA{\textit{Digital Design} \\
\textit{IT University of Copenhagen}\\
Copenhagen, Denmark \\
sebr@itu.dk}

% \IEEEpubid{\begin{minipage}{\textwidth}\ \\[12pt]
% 978-1-7281-4533-4/20/\$31.00 \copyright 2020 IEEE
% \end{minipage}}

\maketitle

\begin{abstract}
Methods for dynamic difficulty adjustment  allow games to be tailored to particular players to maximize their engagement. However, current methods often only modify a limited set of game features such as the difficulty of the opponents, or the availability of resources. Other approaches, such as experience-driven Procedural Content Generation (PCG), can generate complete levels with desired properties such as levels that are neither too hard nor too easy, but require many iterations. This paper presents a method that can generate and search for \emph{complete} levels with a specific target difficulty in only a few trials. This advance is enabled by through an \emph{Intelligent Trial-and-Error algorithm}, originally developed to allow robots to adapt quickly.  
% %In this paper we test the approach on a selection of planning agents. 
Our algorithm first creates a large variety of different levels that vary across predefined dimensions such as leniency or map coverage. The performance of an AI playing agent on these maps  gives a proxy for how difficult the level would be for \emph{another} AI agent (e.g.\ one that employs Monte Carlo Tree Search instead of Greedy Tree Search); using this information, a Bayesian Optimization procedure is deployed, updating the difficulty of the prior map to reflect the ability of the agent. The approach can reliably find levels with a specific target difficulty for a variety of planning agents in only a few trials, while maintaining an understanding of their skill landscape.

\end{abstract}

\begin{IEEEkeywords}
Dynamic Difficulty Adjustment, Intelligent Trial-and-Error, Planning Agents, PCG, MAP-Elites
\end{IEEEkeywords}

\phantom{A} % New line to break the column.

\vspace{-0.7cm} % To remove whitespace.
\section{Introduction}

Dynamic Difficulty Adjustment (DDA) consists of adapting a game online, modifying its difficulty according to the behavior of the player \cite{Zohaib2018}. With DDA, games try to maximize player engagement by continuously presenting challenges that are neither too easy, nor too difficult. If implemented properly, DDA promises to keep players engaged and entertained according to the psychological theory of \textit{flow} \cite{Csikszentmihalyi91}.

%Given its potential, DDA has become an active field of study. 
Methods for DDA range from probability graphs \cite{Xue2017},  Monte Carlo Tree Search and other statistical forward planning algorithms \cite{Demediuk2017, Demediuk2019} to data-driven approaches \cite{Jennings-Teats2010, Zook2012data, Zook2012skill}. However, these approaches come with two main drawbacks. First, they often modulate only a single aspect of a level or game (e.g.\ availability of resources \cite{Hunicke2005},  opponent AI \cite{SILVA2017103}). Second, they do not capture and retain an understanding of what exactly is either causing the player trouble or boring them. 

We develop a variation of the Bayesian-based Intelligent Trial-and-Error Algorithm (IT\&E) to address these two challenges. First developed for fast adaptation in robots \cite{cully2015robots}, this approach has the potential to both serve levels with the ideal level of difficulty and to maintain a player model that is constantly being updated using Bayesian Methods. Another potential advantage of this approach (compared to e.g.\ experience-driven procedural content generation \cite{yannakakis2011experience}) is its ability to find an ideal level faster.

In this paper we study the application of IT\&E for DDA using \emph{different} AI agents as a proxy for \emph{different} human players. AI agents allow for a more clean-cut, controllable environment for testing our approach before applying it to real players. One of our hypotheses is that in a Bayesian optimization process, information on which levels one agent finds difficult can be used as an estimate of how difficult these levels will be for \textit{another} agent. While this paper establishes the main method and shows that it works well for artificial agents, how this approach will scale to human players is an important future research direction.

\section{Background} \label{sec:background}

\subsection{Dynamic Difficulty Adjustment}

The idea behind Dynamic Difficulty Adjustment (DDA) is to modulate the difficulty of a game to keep the player in \textit{flow} \cite{Csikszentmihalyi91}, a psychological state in which a task matches the ability of the user. One of the first approaches to DDA modulated Half-Life's internal economy using probabilistic methods based mostly on Inventory Theory \cite{Hunicke2004}. Here,  the current state of the game is mapped to an adjustment action that affects certain game aspects, such as the availability of health in a level \cite{Hunicke2005}. Other approaches use probabilistic models to optimize player engagement and prevent churn in mobile games \cite{Xue2017}.

For Multi-player Online Battle Arena (MOBA) games, Silva et al. \cite{silva2015adaptive, SILVA2017103} designed different AIs for easy, medium and hard difficulty,  switching between them during deployment based on the performance difference between player and  agent. 

A data-driven approach for a turn-based role-playing game has been proposed by Zook \& Riedl \cite{Zook2012data}, which is based on a tensor factorization method to predict player performance.

Hao et al. \cite{Hao2010PacMan} employed  Monte Carlo Tree Search (MCTS) agents to modulate the difficulty of Pac-Man; by modifying the simulation time of the algorithm, the MCTS agents can be artificially handicapped to perform at different levels of difficulty. Demediuk et al. \cite{Demediuk2017} expand on the idea of using a planning algorithm by introducing variations of MCTS  that change the action selection policy or the heuristic for  evaluating playouts.

However, these aforementioned approaches rely on adapting a limited set of game features to modulate difficulty. While approaches to overcome this limitation exist, allowing  continuously adapting levels for a 2D platformer \cite{Jennings-Teats2010}, they rely on an expensive mass data collection for estimating both level difficulty and player ability. 

Our method evolves sets of entirely new and complete levels that exhibit several  characteristics that can make the level easier or more difficult for different types of players. In contrast to previous approaches, here we maintain a set of evolved levels, updating our estimates of their difficulty. This way we are able to better understand which aspects of the level are challenging for the player and which are too easy (across the different behaviors that are measured).

\subsection{Procedural Content Generation}
The field of Procedural Content Generation (PCG) \cite{shaker2016procedural} is focused on creating game content algorithmically with little or no human intervention (e.g.\ game rules, characters, textures). This approach can benefit players by providing them unique experiences every time they play. Many PCG approaches rely on a fixed set of parameters and randomness to generate content within a heavily constrained space of possibilities, but a recent focus is to apply machine learning approaches to enable a more open-ended generation of content \cite{Summerville2018,risi2019procedural}.

In particular, PCG can be cast as a search problem, which tries to find content -- such as levels -- with particular properties \cite{togelius2011search}. For example, levels can be evolved to have a certain difficulty, to be balanced, and solvable. While this approach has the benefit of being very general and can create content as diverse as game maps \cite{togelius2013controllable}, particle effects for weapons \cite{hastings2009automatic} or even game rules \cite{togelius2008experiment}, searching for these artifacts can take many iterations. To limit the space of content the algorithm has to search through, these approaches can also be combined with self-supervised learning. \emph{Latent Variable Evolution} (LVE)  \cite{bontrager2018deepmasterprints} is one such approach that has been applied successfully to creating artificial fingerprints \cite{bontrager2018deepmasterprints} and Super Mario Bros. levels \cite{volz2018evolving}. In LVE,  a generator is trained on existing content and evolution only searches the space of latent variables given as input to the pre-trained generator. However, even in this case, search typically still takes many generations \cite{volz2018evolving}. 

In contrast to existing search-based PCG work, the proposed approach can find levels with the right difficulty in only a few trials, by incorporating useful priors computed beforehand and using a sample efficient search. The algorithm is tested on generating new levels described in the Video Game Description Language (VGDL), which is explained next.

\subsection{The GVGAI Framework}

The General Video Game AI (GVGAI) framework provides a set of tools for testing artificial agents \cite{Perez-Liebana2016}. It was created to evaluate how well agents generalize over multiple games, which is the focus of the multiple competitions that are being held since its inception. GVGAI allows the user to easily generate and run games specified in  the Video Game Description Langauge (VGDL) \cite{schaul2013pyvgdl}. 

The IT\&E for games approach introduced in this paper creates levels for a \textit{Zelda}-like video game. The objective in this game consists of picking up a key and navigating towards the final goal. The level is arranged as a dungeon with enemies of three different kinds that move randomly at different speeds. The player loses the game if an enemy touches their avatar. The player is also equipped with a sword to kill opponents, which gives additional points.

The GVGAI framework comes with several game-playing agents that implement different planning algorithms. In this paper we test the algorithm to quickly find levels with the right difficulty for the following eight agents:
\begin{itemize}
    \item \textbf{DoNothing}, an agent that stands still.
    \item \textbf{Random}, performs a random action at each timestep.
    % \item \textbf{One Step Lookahead} (OSLA), an agent that chooses the best next action according to a simple score heuristic\footnote{expand on this heuristic (?)} that assesses the next states.\todo{add description of heurstic}
    \item \textbf{One Step Look-ahead} (OSLA), chooses the best next action according to a simple score heuristic that assesses the next states and picks the best performing one. This heuristic encourages the agent to kill opponents if they are one step away in the tree and to move towards whatever maximizes score if they are also at a one-step distance.
    \item \textbf{Greedy Tree Search} (GTS),  searches the game tree for the next best state using the score as a heuristic. The search is limited by a time budget of 40ms per step imposed by the framework itself. 
    \item \textbf{Random Search} (RS), creates a set of random playtraces from the current state up to a certain depth in the game tree (as much as the framework time budget allows), and follows the first step of the best playtrace according to a state heuristic that returns either the score or $\pm10^6$ if the agent wins or loses respectively.
    \item \textbf{Rolling Horizon Evolution} (RHEA) \cite{Perez2013RHEA} uses an evolutionary algorithm to compute an optimal playtrace up to a certain depth (with time budget of 40ms). The fitness function follows the same heuristic as in the RS setup. 
    \item A vanilla implementation of the \textbf{Monte Carlo Tree Search} (MCTS) algorithm \cite{gvgaibook2019}. MCTS constructs an estimate of the average value of nodes in the game tree by running rollouts up to the end of the game. This value is guided by the same score heuristic that governs RS and RHEA, and the action selection takes into account an extra additive term that encourages exploration.
    \item \textbf{Open Loop Expectimax Tree Search} (OLETS), the agent that won the first edition of the single-player GVGAI competition \cite{gvgaibook2019}. At each step, OLETS runs a simulation in which it assesses the value of next actions according to the \textit{Open Loop Expectimax} heuristic that includes not only the average value of the node but also the maximum value among its children. \cite{Perez-Liebana2016GVGAIagents}. 
\end{itemize}

\begin{lstfloat}
\begin{lstlisting}
procedure MAP-Elites(n_iters, n_init):
    $\mathcal{P}$ = $\varnothing$
    $\mathcal{X}$ = $\varnothing$
    for iter = 1 $\to$ n_iters:
        if iter < n_init:
            $x'$ = random_solution()
        else:
            $x$ = random_selection($\mathcal{X}$)
            $x'$ = random_variation($x$)
        $b'$ = behavior_descriptor($x'$)
        $p'$ = performance($x'$)
        if $\mathcal{P}(b') = \varnothing$ or $\mathcal{P}(b') < p'$:
            // update the elite in the cell
            $\mathcal{P}(b') = p'$
            $\mathcal{X}(b') = x'$
    return $(\mathcal{P}, \mathcal{X})$ // behavior-performance map
\end{lstlisting}
\caption{MAP-Elites' pseudocode.}
\label{code:MAP-elites}
\end{lstfloat}

\subsection{Illumination Algorithms}

While the goal of a typical evolutionary algorithm is to produce one solution that maximizes a given performance metric, the goal of illumination algorithms is to find high-performing solutions in different sections of the search space \cite{Mouret2015,Lehman2011NoveltySearch}. By extending the search, illumination algorithms can find a plethora of individuals with high fitness that have different characteristics.

In this paper, we use \textit{MAP-Elites} \cite{Mouret2015} (Algorithm~\ref{code:MAP-elites}). In MAP-Elites, the objective consists of obtaining an archive of elites, each expressing a different \textit{behavior} in a low dimensional space. This behavior space is divided into cells, and each cell is identified by its centroid. Each cell maintains the best performing individual whose behavior is closest to its centroid. Normally, individuals are chosen uniformly at random from the current elites to be mutated. In the work here, MAP-Elites produces a map of levels that vary across certain dimensions such as the space covered or the distance to the goal.

Variants of the MAP-Elites algorithm have been applied to many game tasks before, such as evolving agents for game balancing in Hearthstone \cite{Fontaine2019}, mixed-initiative co-creation tools for dungeon level design \cite{Alvarez2019}, or level creation for Bullet Hell games \cite{Khalifa2018}. To the best of our knowledge, our approach is the first one to leverage the combination of MAP-Elites and Bayesian Optimization (known as the Intelligent Trial-and-Error algorithm) for fast game difficulty adjustment.

\subsection{Intelligent Trial-and-Error} \label{subsec:ITAE_background}

Finding the optimal level of difficulty for a player can be thought of as an optimization process. Bayesian Optimization (BO) consists of learning a regression model for the function to be optimized and using it to drive the optimization \cite{BayesOptSurvey}. The core idea of BO is to maintain a prior over all the possible forms of the objective function, and update it after querying using Bayes' rule. Bayesian regression is well known for being data-efficient and for working even for black-box functions.

The Intelligent Trial-and-Error algorithm (IT\&E) \cite{cully2015robots} is a form of BO that relies on Gaussian Process Regression \cite{Rasmussen2004} and the MAP-Elites algorithm \cite{Mouret2015}. In Gaussian Process Regression, the objective is to learn a function $f(x)$ using an assumed prior function $\mu_0(x)$ and a kernel function $k(x, x')$ that describe the mean and covariance of the process respectively (we write $f(x) \sim \mbox{GP}(\mu_0(x), k(x, x'))$.  A common choice for a kernel function $k$ (which we will use for our experiments), is the Matérn$_{5/2}$ kernel, given by
\begin{equation}\label{eq:matern}
    k(x,x';\sigma) = \sigma^2 \left(1 + \sqrt{5} r + \frac{5}{3}r^2\right) \exp\left(- \sqrt{5} r\right)
\end{equation}
where $r$ is the distance between $x$ and $x'$, and $\sigma$ is a scalar hyperparameter \cite{gpy2014}. By \textit{conditioning} on newly arrived data $\bm{x}=[x_i]_{i=1}^t$ and $\bm{f}=[f_i]_{i=1}^t$, our estimation of $f(x)$ can be iteratively updated. If we want to predict $\tilde{f}$ for a particular $\tilde{x}$, we condition on $\bm{x},\bm{f}$ and $\tilde{x}$ to get a normal distribution for $\tilde{f}$:
\begin{equation}\label{eq:GP_update}
\begin{aligned}
    \tilde{f}\,|\,\tilde{x},\bm{x},\bm{f} \sim \mathcal{N}(&\mu_0(x) + \bm{k}(\tilde{x})^TK^{-1}\bm{f},\\
    &k(\tilde{x},\tilde{x}) -\bm{k}(\tilde{x})^TK^{-1}\bm{k}(\tilde{x})),
\end{aligned}
\end{equation}
where $\bm{k}(\tilde{x})=[k(\tilde{x}, x_i)]_{i=1}^t$, $K = [k(x_i, x_j)]_{i,j=1}^{t} + \sigma_{\text{noise}}^2I$, and $\sigma_{\text{noise}}$ is a hyperparameter.

% MIGUEL: COULD BE OMITTED: "This is equal to sampling at the upper confidence bound."
% Using Gaussian Processes, the IT\&E algorithm updates a behavior map obtained via MAP-Elites to approximate the real performances. Let $\mathcal{P}(x)$ be the behavior-performance map (see Algorithm~\ref{code:MAP-elites}). The algorithm starts by assigning $\mu_0(x) = \mathcal{P}(x)$, then, the algorithm selects and deploys an elite $x_{t+1}\in\mathcal{X}$ by maximizing $\mu_t(x_{t+1}) + \beta\sigma_t(x_{t+1})$ where $\sigma_t(x_{t+1})$ is the posterior standard deviation. Once the real performance $p_{t+1}$ is obtained, a new approximation of the performance map is computed by using eq. \ref{eq:GP_update}.

The IT\&E algorithm starts by computing a prior map using MAP-Elites, and then uses Gaussian Processes to update these beliefs about the objective function. Let $\mathcal{P}(x)$ be the behavior-performance map (Algorithm~\ref{code:MAP-elites}). 
%Commented this out. Doesn't seem to add much: e.g.\ levels $x$ and their measured difficulty $\mathcal{P}(x)$ for an agent)
%, e.g.\ $x$ a level and $\mathcal{P}(x)$ its estimated difficulty). 
The algorithm first assigns $\mu_0(x) = \mathcal{P}(x)$, and then selects and deploys an elite $x_{t+1}\in\mathcal{X}$ by maximizing $\mu_t(x_{t+1}) + \beta\sigma_t(x_{t+1})$ where $\sigma_t(x_{t+1})$ is the posterior standard deviation. Once the real performance $p_{t+1}$ (i.e.\ how difficult the map is for a new agent) 
%(i.e.\ how well thi of level $x_{t+1}$) 
is obtained, a new approximation of the performance map is computed with Equation~\ref{eq:GP_update}. In other words, the algorithm starts by sampling the best performing $x$ from the prior map constructed using MAP-Elites, and updates this map once the actual performance of $x$ is received.

\section{Approach: Fast Game Difficulty Adjustment through IT\&E}
\begin{figure}[t]
    \centering
    \includegraphics[width=1\columnwidth]{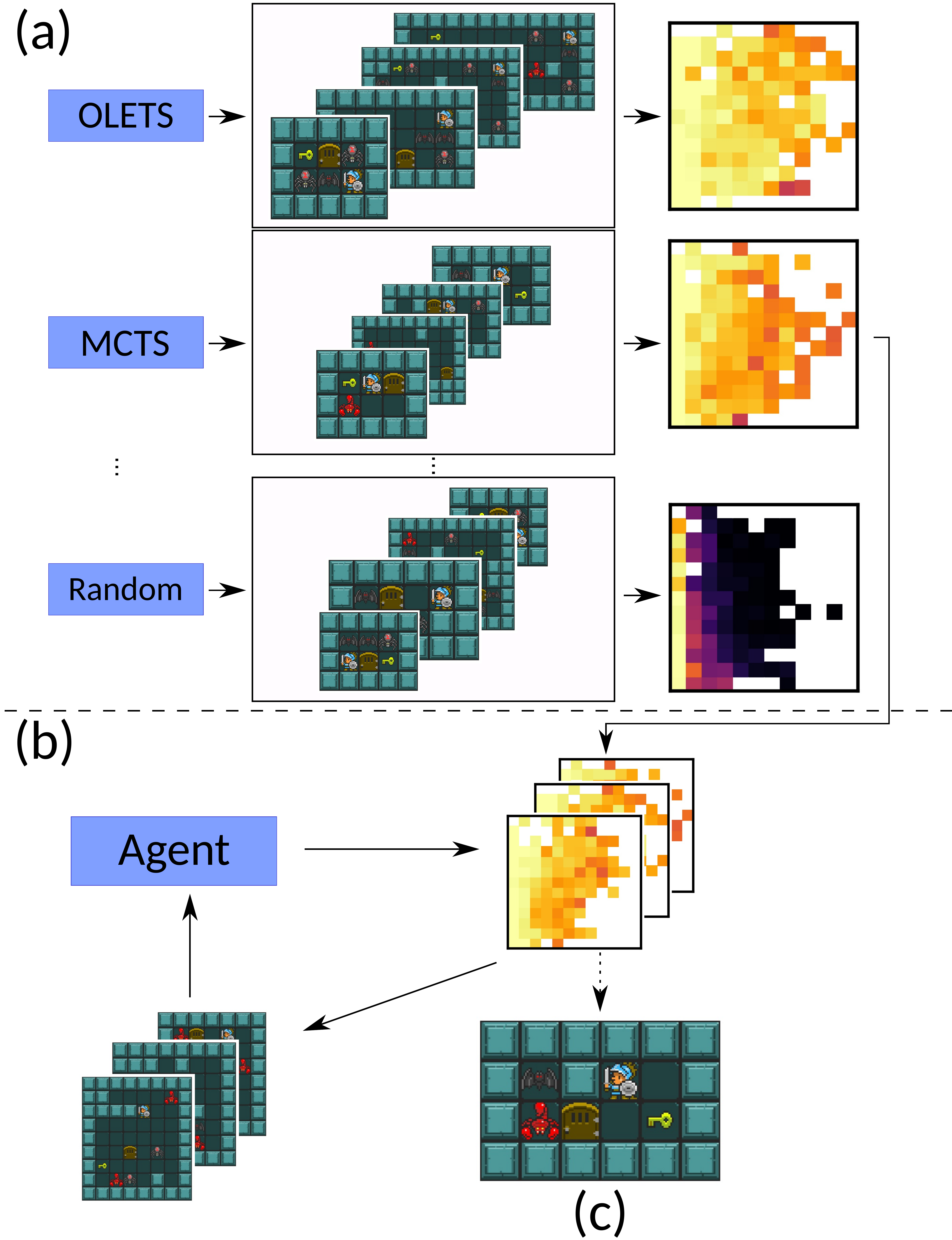}
    \caption{\textbf{An overview of our IT\&E level generation approach}. (a) First, a diverse set of levels are evolved such that their difficulty, evaluated as the win rate of several AI agents, is optimal. This set of diverse levels is organized in a map such that levels that have similar \textit{behavioral} features (e.g. same amount of enemies, or level coverage) are close. Each cell in the map corresponds to a level, and the color represents how easy (bright) or difficult (dark) a level is for said agent. (b) These level difficulty maps are then used as priors in a Bayesian optimization procedure, which iterates over levels to quickly select an optimal level for \emph{another} AI agent (c).}
    \label{fig:overview}

\end{figure}

In this paper, we use a planning agent $P'$, as a proxy for a human player. We break down the task of quickly finding optimally difficult levels for $P'$ into the following steps (Fig.~\ref{fig:overview}): (\textbf{1}) Evolve a diverse set of levels with near-optimal win-rates for several AI agents (excluding $P'$) using the MAP-Elites algorithm (Fig.~\ref{fig:overview}a). Note: this step, although expensive, only needs to be done once and can be done offline. (\textbf{2})  Use the IT\&E algorithm to quickly find an optimal level for $P'$ (Fig.~\ref{fig:overview}c), using the performance of \emph{other} AI agents as a prior estimate of the performance of $P'$ on each level (Fig.~\ref{fig:overview}b).

In our implementation of IT\&E, this process stops after finding an individual whose performance is above a certain predefined bound. The original IT\&E implementation only stops after finding an individual whose performance is above $\alpha\max(\mu_t(x))$ for a hyperparameter $\alpha\in[0,1]$, that is, when our estimates of the real performances indicate that there does not exist a better performing individual to be tested (up to a certain percentage governed by $\alpha$).

\subsection{Generating diverse and optimally difficult levels} \label{subsec:MAP_Elites_exp}

% \begin{table}
%     \centering
%     \resizebox{\columnwidth}{!}{
%     \begin{tabular}{l|lllll|r}
% \toprule
% {} & Easy & \multicolumn{3}{c}{Medium} & Hard & {} \\
% Agent & $1 \geq w \geq0.8$ & $0.8 > w \geq0.6$ & $0.6 >w \geq 0.4$ & $0.4 > w \geq 0.2$ & $0.2 > w \geq 0$ &  Total \\
% \midrule
% OLETS                  &  326 &           2 &      1 &           0 &    0 &     329 \\
% MCTS             &  319 &          30 &      5 &           2 &    0 &     356 \\
% RHEA             &  246 &          82 &     13 &           1 &    0 &     342 \\
% RS               &  268 &          53 &      6 &           1 &    0 &     328 \\
% GTS       &  111 &          79 &     69 &          39 &   34 &     332 \\
% OSLA &   60 &          17 &     22 &          14 &  220 &     333 \\
% Random           &   48 &           9 &     33 &          41 &  191 &     322 \\
% doNothing              &    0 &           0 &      0 &           0 &  341 &     341 \\
% \bottomrule
% \end{tabular}
% }
%     \caption{ \textbf{Amount of levels per difficulty}: Levels with win rate in e.g. [0.8, 1] are easy for the agent, while levels in [0, 0.2) can be considered hard. The \textit{advanced} agents find most levels easy; the map computed using greedyTreeSearch contains levels of all difficulties, and the simpler controllers find most levels too difficult.} % TODO: change the column names to "easy, medium easy..." (?)
%     \label{tab:level_count}
% \end{table}

\begin{table}
    \centering
    \resizebox{\columnwidth}{!}{
    \begin{tabular}{l|c|ccc|c}
\toprule
{} & \multicolumn{1}{c}{Easy} & \phantom{123} & Medium & {} & \multicolumn{1}{c}{Hard}\\
Agent & $1 \geq w \geq0.8$ & $0.8 > w \geq0.6$ & $0.6 >w \geq 0.4$ & $0.4 > w \geq 0.2$ & $0.2 > w \geq 0$\\
\midrule
OLETS                  &  326 &           2 &      1 &           0 &    0 \\
MCTS             &  319 &          30 &      5 &           2 &    0 \\
RHEA             &  246 &          82 &     13 &           1 &    0\\
RS               &  268 &          53 &      6 &           1 &    0\\
GTS       &  111 &          79 &     69 &          39 &   34 \\
OSLA &   60 &          17 &     22 &          14 &  220\\
Random           &   48 &           9 &     33 &          41 &  191 \\
doNothing              &    0 &           0 &      0 &           0 &  341 \\
\bottomrule
\end{tabular}
}
    \caption{ \textbf{Amount of levels per difficulty}: Levels with win rate between 0.8 and 1 are easy for the agent, while levels with win rate between 0 and 0.2 can be considered hard. The \textit{advanced} agents (OLETS, MCTS, RHEA and RS) find most levels easy; the map computed using Greedy Tree Search contains the most variety of difficulties, and the simpler controllers find most levels too difficult.} % TODO: change the column names to "easy, medium easy..." (?)
    \label{tab:level_count}
\end{table}

Before we can run IT\&E to quickly find a level with the right difficulty for a new agent (Section~\ref{subsec:ITAE_exp}), we first have to  create initial level maps for all the different AI-playing agents (Fig.~\ref{fig:overview}a) through MAP-Elites (Algorithm~\ref{code:MAP-elites}), which is detailed in this section. The algorithm requires two functions:  \texttt{random\_solution()} which returns a random level and \texttt{random\_variation(level)} which randomly mutates a level.
\captionsetup{labelfont=md}
\begin{figure}[!t]
    \centering
    \begin{subfigure}[t]{.122\textwidth}
      \centering
      \includegraphics[width=1\linewidth]{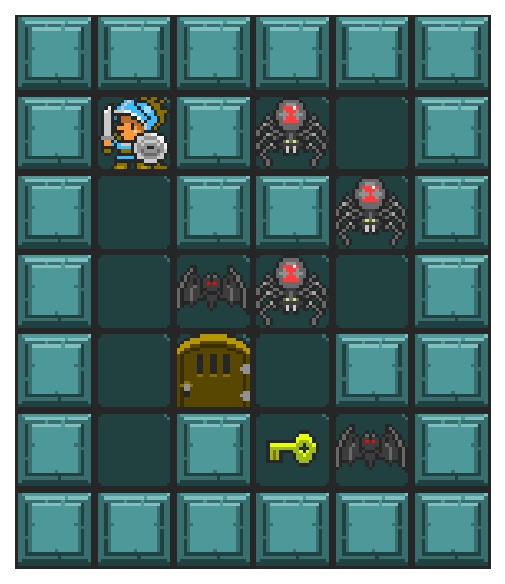}
      \caption{}
      \label{fig:sub_original}
    \end{subfigure}%
    \begin{subfigure}[t]{.122\textwidth}
      \centering
      \includegraphics[width=1\linewidth]{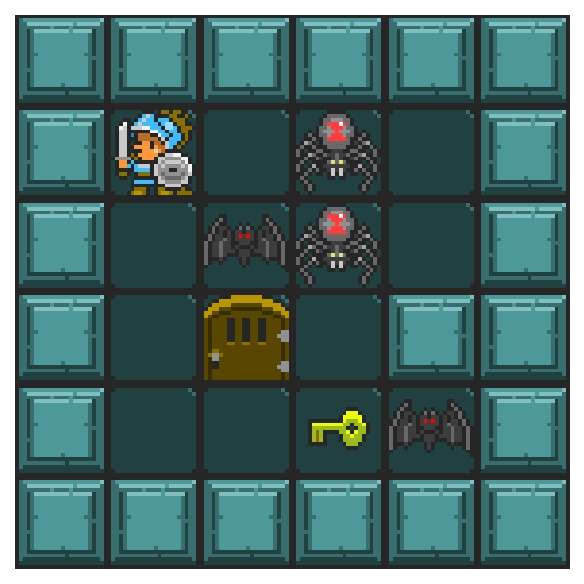}
      \caption{}
      \label{fig:sub_var1}
    \end{subfigure}%
    \begin{subfigure}[t]{.122\textwidth}
      \centering
      \includegraphics[width=1\linewidth]{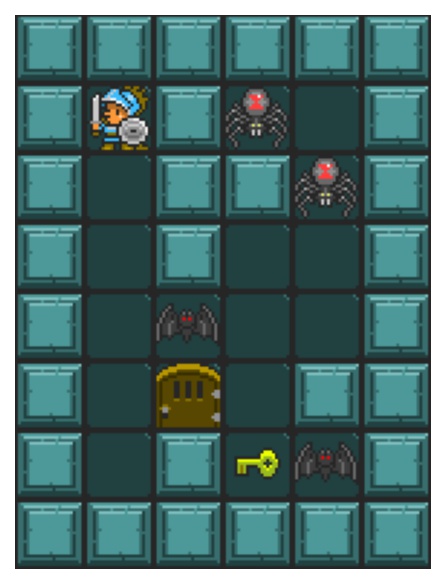}
      \caption{}
      \label{fig:sub_var2}
    \end{subfigure}%
    \begin{subfigure}[t]{.122\textwidth}
      \centering
      \includegraphics[width=1\linewidth]{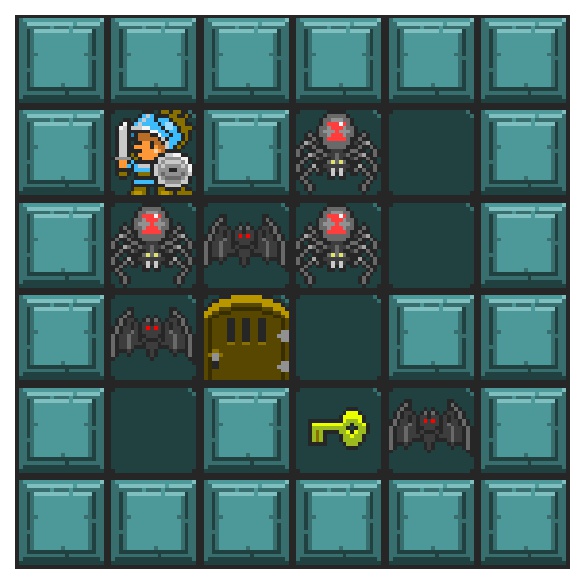}
      \caption{}
      \label{fig:sub_var3}
    \end{subfigure}
    \captionsetup{labelfont=bf}
    \caption{ \textbf{Level creation and mutation}: Shown are a generated example level (a) and three different level mutations (b--d). In (b) the third row and two walls were removed; in (c) a new row and a wall were added, and one enemy was removed; in (d) the third row was removed and two enemies were added.}
    \label{fig:random_variations}
\end{figure}
\captionsetup{labelfont=bf}
The function \texttt{random\_solution()} is implemented as follows: 
\begin{enumerate}
\item Sample the width $w$ and height $h$ at random between 3 and 9.
\item Sample the amount of enemies $e$ at random between $\lfloor\min(w, h)/2\rfloor$ and $\min(w, h)$.
\item If $\min(w, h) > 3$, sample a random integer $i$, similarly to $e$, for the amount of inner walls in the level.\footnote{Notice that if $\min(w, h) = 3$, then any inner wall would almost surely block the path between the avatar, the key, and the goal, making the game unwinnable.}
\item If there is not enough room for placing the player, key, goal, enemies and walls, i.e. $i+e+3 > (w-2)(h-2)$, adjust $h = h + 1$ or $w = w + 1$ (selecting at random) until there is enough room. % rbp: couldn't you sample i from $[0; wh-(3+e)]$ instead. That way you'd be sure $i+e+3 \leq wh$?, mgd: 
\item Create a level with width $w$ and height $h$ with walls at the borders. % 
\item Place the avatar, key, and goal at random in the level. 
\item Repeating $e$ times: randomly select a type of enemy and place it in a random available position.
\item Compute the A* path between the avatar and the key, and between the key and the goal. Mark the positions in this path as occupied.
\item If $i$ is the number of inner walls and $a$ is the number of available positions after removing the positions of the paths, then place $\min(a, i)$ inner walls at random.
\end{enumerate}

The function \texttt{random\_solution()} always returns a solvable level, because the A* path between avatar, key and goal are preserved. To mutate these levels, the function  \texttt{random\_variations(level)} takes a level and performs the following steps: 
\begin{enumerate}
\item Expands or contracts the width/height by adding or removing one column/row at random (only considering the columns/rows that don't have either the avatar, key nor goal) if possible.
\item Adds or removes random enemies by sampling a random integer between -2 and 2, where negative numbers imply removing instead of adding.
\item Adds or removes inner walls similar to the previous step, verifying that the connectivity between Avatar, key, and goal is not broken.
\end{enumerate}

With the two functions \texttt{random\_solution()} and \texttt{random\_variation(level)} we can create new levels and mutate them (Fig.~\ref{fig:random_variations}).
\begin{figure}
    \centering
    \includegraphics[width=0.8\columnwidth]{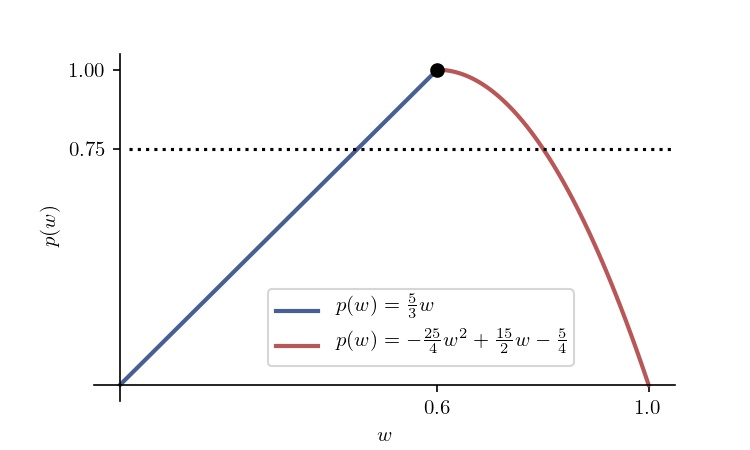}
    % MIGUEL: COULD BE OMITTED: the formula.
    \caption{\textbf{Performance function}: since using win rate alone would encourage the creation of only easy levels, we used a custom performance function that has its maximum at 60\% win rate. The rationale for choosing 60\% win rate is that easy levels (with win rates closer to 100\%) would be boring for the player, and levels that are too difficult (with win rates closer to 0\%) would be frustrating. This function is defined as $p(w) = (5/3)w$ up to $0.6$ and as $p(w) = -(25/4)w^2 + (15/2)w - 5/4$ from $0.6$ to $1$. When computing the ideal level for an agent using a different prior (see Subsec.~\ref{subsec:ITAE_exp}), we stopped after finding a level such that $p(w) \geq 0.75$.}
    \label{fig:performance_function}
\end{figure}

The performance function should drive MAP-Elites towards creating a map of levels that are neither too easy nor too hard for the particular AI agent.  Choosing the win rate itself as the value to be optimized would drive the process towards creating maps with levels that are too easy (i.e.\ levels in which the win rate is 100\%). 
%To be able to find optimally difficult levels, 
Therefore we define a performance metric $p$ that has a maximum of $1$ when the agent performs at win rates of 60\% and goes down to 0 at both 0\% and 100\% win rate (Fig.~\ref{fig:performance_function}). With this custom fitness function, we are encouraging the creation of levels with 60\% win rate. Performance for each agent deployed in a level is measured as the average across 40 rollouts. 
%In each iteration of the MAP-Elites algorithm, we deploy an agent in the level for 40 rollouts, and use the average performance across
%nd use the win rate after these rollouts to measure how difficult the agent finds the level.
For each generated level, we compute the following behavioral features that determined its location in the map:
\begin{itemize}
    \item \textbf{Space coverage:} the percentage of the level that is filled with prompts (how \textit{crowded} the level is).
    \item \textbf{Leniency:} the number of enemies in the level.
    \item \textbf{Reachability:} the sum of distances of the paths between the avatar and the key and the key and the goal.
\end{itemize}

To summarise, at the end of the map generation process we have different maps of elites for the different agents, together with average win rates for each level.

\section{Results}
\begin{figure}[t]
    \centering
    \includegraphics[width=1\columnwidth]{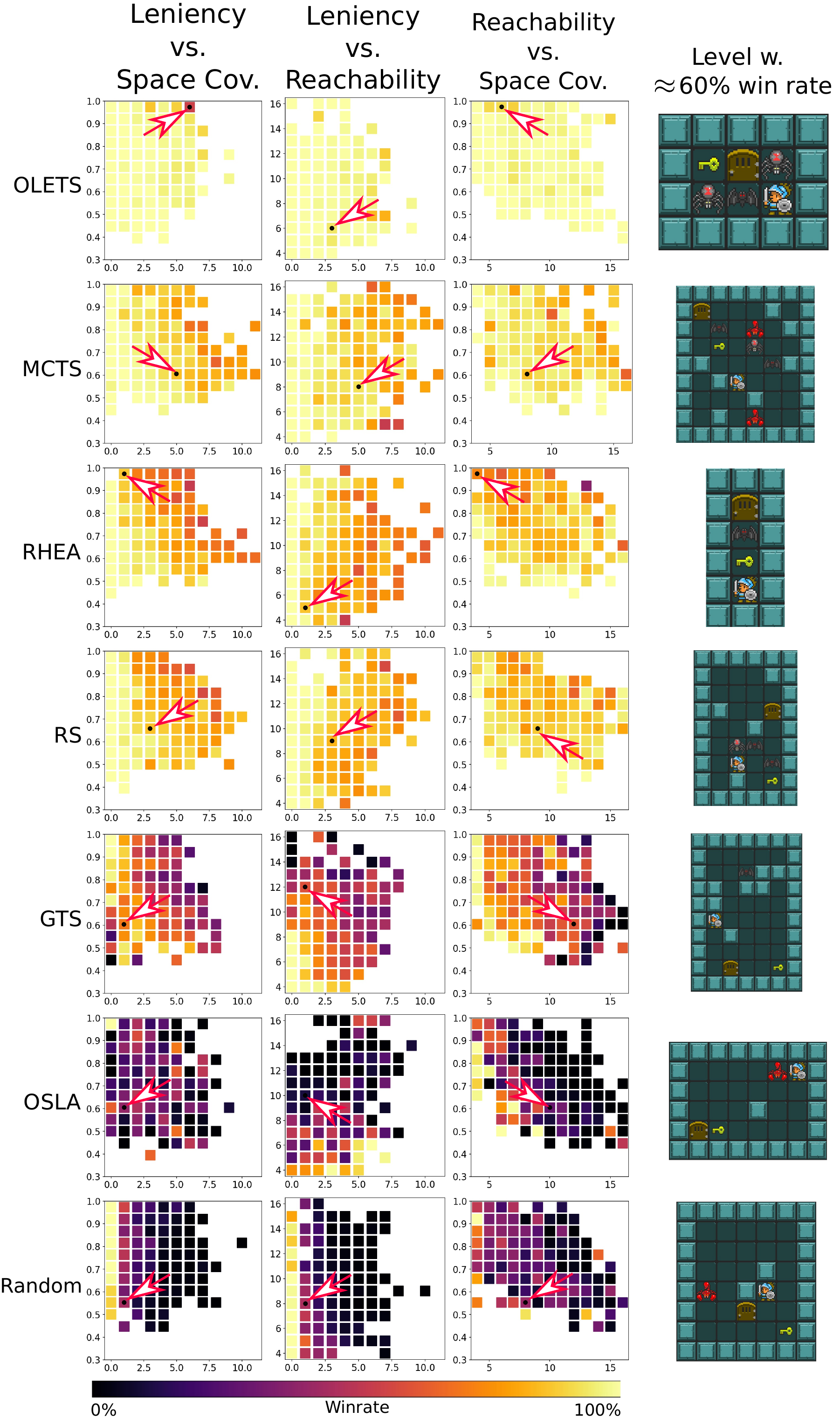}
    \caption{\textbf{Archive of elites for different agents}: Shown are the final generation of the behavior map obtained for different planning agents, colored by win rate. The brighter the color, the higher the win rate for the agent. The position of a level with a particular win rate (60\%) is highlighted in the map. These maps illustrate what each of the agents finds easy or difficult. For example, the OSLA agent struggles at levels with higher reachability (i.e.\ longer distances to the key and goal), because it only looks at the next nodes in the game tree.}
    \label{fig:priors}
\end{figure}

For each agent, MAP-Elites was run for 10 generations, with an initialization of 100 levels and with 50 iterations per generation after that (Fig.~\ref{fig:priors}). Since the behavioral features are computed in $\mathbb{R}^3$,  maps show a 2D projection obtained by averaging over the remaining behavioral feature. An example of an evolved level with a win rate of approximately 60\% is also shown.

% New (Miguel, 27/03/2020 - 17:00)
These maps reflect the aptitudes and deficiencies of each of the different agents employed in the experiments. The OLETS agent finds most levels too easy (for the behavioral features that were measured). This observation is supported by the results in Table~\ref{tab:level_count}, which show a count of levels, segmented by difficulty. This is also the case for most of the \textit{advanced} planning agents we investigated (MCTS, RHEA and RS).

\begin{table*}[]
    \centering
    \resizebox{2\columnwidth}{!}{
\begin{tabular}{lrrrrrrr}
\toprule
{} &                  OLETS &                   MCTS &                   RHEA &                     RS &          GTS &                   OSLA &                 Random \\
Prior\textbackslash Agent      &                        &                        &                        &                        &              &                        &                        \\
\midrule
OLETS            &            1.3 (10/10) &            \phantom{0}\textbf{1.2 (10/10)} &            1.4 (10/10) &            1.6 (10/10) &   7.1 (9/10) &  \phantom{00.0}(0/10) &  \phantom{00.0 }(0/10) \\
MCTS             &            11.7 (7/10) &            \phantom{0}1.2 (10/10) &            2.1 (10/10) &            2.9 (10/10) &  6.7 (10/10) &           11.7 (10/10) &  \phantom{00.0 }(0/10) \\
RHEA             &  \phantom{00.0 }(0/10) &            \phantom{0}2.5 (10/10) &            1.1 (10/10) &            \textbf{1.0 (10/10)} &  3.2 (10/10) &            15.6 (7/10) &  \phantom{00.0 }(0/10) \\
RS               &            \phantom{0}\textbf{5.4 (8/10)} &            \phantom{0}2.0 (10/10) &            \textbf{1.1 (10/10)} &            1.5 (10/10) &  3.7 (10/10) &           12.0 (10/10) &  \phantom{00.0 }(0/10) \\
GTS              &  \phantom{00.0 }(0/10) &            \phantom{0}11.6 (9/10) &            7.1 (10/10) &             7.5 (8/10) &  1.1 (10/10) &            11.7 (3/10) &            2.6 (10/10) \\
OSLA             &  \phantom{00.0 }(0/10) &           10.5 (10/10) &            3.8 (10/10) &            5.8 (10/10) &  5.8 (10/10) &            1.2 (10/10) &            20.0 (2/10) \\
Random           &  \phantom{00.0 }(0/10) &  \phantom{0}\phantom{00.0 }(0/10) &  \phantom{00.0 }(0/10) &  \phantom{00.0 }(0/10) &  13.7 (3/10) &            \textbf{3.5 (10/10)} &            1.3 (10/10) \\
doNothing        &  \phantom{00.0 }(0/10) &             \phantom{0}\phantom{0}8.8 (6/10) &             3.0 (1/10) &            12.0 (3/10) &  \textbf{2.4 (10/10)} &             9.5 (2/10) &            11.4 (5/10) \\
Baseline (noise) &  \phantom{00.0 }(0/10) &            \phantom{0}15.7 (7/10) &            3.9 (10/10) &            4.5 (10/10) &  \textbf{1.0 (10/10)} &  \phantom{00.0 }(0/10) &            \textbf{2.0 (10/10)} \\
\bottomrule
\end{tabular}
}
    %\caption{\textbf{Mean iterations to find a level with ideal difficulty.} This table presents the average amount of iterations to find a compensatory level for all pairs of priors and agents. It also presents the amount of successful adaptions (e.g. 7/10 means that the algorithm was able to find a level with win rate in [0.5, 0.7] in 7 out of 10 runs). This average was computed only on successful runs. The main diagonal in this table represents deploying an agent with a prior that has been constructed, so it is to be expected to find a compensatory level in few tries. The main insights of this table lie outside this diagonal. These results show that we are able to find an ideal level of difficulty for an agent using the prior of \textit{another} after few updates.}
    \caption{\textbf{Mean iterations to find a level with ideal difficulty.} This table presents the average number of iterations to find a compensatory level for all pairs of priors and agents. It also presents the amount of successful adaptions (e.g. 7/10 means that the algorithm was able to find a level with win rate in approximately [0.5, 0.7] in 7 out of 10 runs). This average was computed only on successful runs. These results show that we can find an ideal level of difficulty for an agent using the prior of \textit{another} after a few updates. There is a relationship between the skill of the agent and the quality of the map (i.e.\ the skill of the agent that was used to create it). Deploying IT\&E using the map of a skilled agent on a more basic one translates to more updates, and vice versa. This makes intuitive sense: the levels in the archive map of a skilled agent might be \textit{too difficult} for a more basic agent, and it takes more iterations to find a level that is \textit{easy enough}.}
    \label{tab:iterations}
\end{table*}

On the other hand, the Random agent only finds agents without opponents easy, since it will eventually stumble upon the key and the goal. From the basic agents, Greedy Tree Search (GTS) is better able to deal with sparsity and longer distances to the goals. This result can be explained by the fact that GTS can search deeper into the game tree. The exact opposite is true for the One Step Look-Ahead agent (OSLA), which can only deal with levels in which the objectives are close-by (low \textit{reachability}). The presence or absence of enemies (measured by \textit{leniency}) has no impact on the performance of OSLA.

%In the case of OSLA, the presence of enemies seem to help it navigate and explore the level more, given its\todo{it's? it performs a shallow search?} shallow search on the game tree.
% \todo{find the goal?} MIGUEL: yes.

These insights are also reflected by the high-performing levels (i.e.\ win rate close to 60\%; see  Fig.~\ref{fig:performance_function}) shown in Fig.~\ref{fig:priors}. OLETS successfully navigates levels even when surrounded by enemies, and the lower performing agents can only solve levels with fewer enemies. GTS can navigate to further goals (e.g.\ the key on the lower right), while the Random agent struggles even when it is close to the objectives.

\subsection{Dynamic Difficulty Adjustment through IT\&E} \label{subsec:ITAE_exp}

After computing all the elite archives, we test whether the IT\&E can quickly provide a level of appropriate difficulty if the agent changed starting from a particular performance map. For example, would the priors of a map created through an MCTS agent allow faster adaptation for an OLETS agent than priors from a DoNothing agent? 

We deploy a variation of the original IT\&E algorithm \cite{cully2015robots} that stops after finding a compensatory level with win rate approximately between 0.5 and 0.7. Expressing this win rate in terms of performance function (Fig.~\ref{fig:performance_function}), means finding a level   with performance $p \geq 0.75$.

 The experiment of searching for the optimal level was repeated ten times for each pair of agents. Since the goal is to test for fast adaption, each experiment was stopped if the algorithm did not find a compensatory level in the first 20 updates. In that case, the search is counted as unsuccessful. For our experiments, we choose a Matérn$_{5/2}$ kernel with lengthscale $\sigma=1$ (Eq.~\ref{eq:matern}), noise variance given by $\sigma_{\text{noise}}^2=0.1$, and $\beta=0.03$ (Sec.~\ref{subsec:ITAE_background}).

In addition to the agents discussed in the previous section, a baseline behavior map is constructed by assigning random performances to the cells of the DoNothing prior (i.e.\ a random selection of levels). Running an experiment with this baseline is akin to  choosing a random ordering off all the levels in the archive of the DoNothing agent, starting with the first one, and running continuous updates thereafter. With this baseline, we aim to measure whether the priors are useful for quick adaptation, compared to just sampling levels at random.

Table~\ref{tab:iterations} shows the mean number of iterations it took  IT\&E  to find a level with the right amount of difficulty for  all successful runs, together with the number of successful runs. Given the prior of \textit{another} agent, the algorithm can find a level in only a few iterations (at most 16) for all agents. The skill (or lack thereof) of certain agents make it difficult to find a compensatory level: the OLETS agent, in particular, performs at such a high level that the priors of more basic agents have no level difficult enough; the opposite is true for the random agent, which performs so poorly that the advanced agent priors have no easy enough levels.
%  \todo{for most agents?}

% \todo{Update this description if we're sticking with this banner image.}; MIGUEL: done.
% \todo{discuss variance in beh. features}; MIGUEL: done.
Fig.~\ref{fig:updates} shows an example of the IT\&E adaptation steps. Starting with the Random agent archive as a prior, the algorithm finds a level that is difficult enough for the One Step Look-Ahead (OSLA) agent after 3 updates. The heatmap shows performance $p(w)$ instead of win rate, and brighter colors represent levels that are close to the ideal difficulty of 60\%. The first level selected is the best performing one for the Random agent (Fig.~\ref{fig:priors}), and the OSLA agent loses all 40 rollouts. Since this level is too difficult, the search is moved towards a level in which the key and goal are one step away from each other, achieving 100\% win rate. Because the stopping criterion has not been achieved, the algorithm continues and in the next iteration finds a level that is quite similar to the ideal level in the OSLA prior (Fig.~\ref{fig:priors}). This level has approximately the ideal difficulty for OSLA since it performs at 50\% win rate. Since color represents how close a level is to optimal difficulty, it is worth asking why the maps seem to select levels in cells that are not the brightest. This can be explained by two reasons: the acquisition function in the Gaussian Process encourages exploration (Sec.~\ref{subsec:ITAE_background}), and, since these maps are a projection into two dimensions obtained by averaging over the remaining behavioral feature, the behavior selected may be the highest performing one. This last fact also explains why the changes in the map seem to be small when they actually are not.

A surprising result in these experiments is that the baseline prior (which consists of a random assignment of performance to the levels in the doNothing prior) serves as a good prior for the IT\&E algorithm in most agents. We argue that this is happening because of two reasons. First, in this random ordering of levels in the baseline prior, the best performing levels coincide by chance with the ideal levels for Greedy Tree Search and the Random agent; secondly, what makes the IT\&E algorithm work lies more in the Bayesian update component than in the quality of the map. This last fact has also been discussed in other applications of the IT\&E algorithm \cite{Tarapore2016}. However, using a good prior still leads to faster difficulty adjustment in most cases.

\section{Discussion and Conclusion}

In this article we tested whether the Intelligent Trial-and-Error algorithm (IT\&E) \cite{cully2015robots} could be used for Dynamic Difficulty Adjustment (DDA); for this, we evolved several archives of difficult levels for different planning agents and lay them in a \textit{behavior map} using MAP-Elites \cite{Mouret2015}, and we used these maps as priors in a Gaussian Process. We were able to find levels with ideal difficulty for an agent using the archive of \textit{another} agent after a few updates.

% The results we got (first half)
These behavior maps reflect the aptitudes and deficiencies of the agents that drove their evolution. The Random agent's archive shows better performance when there are no enemies in the level, while higher-performing planning agents such as Monte Carlo Tree Search (MCTS) are able to deal with opponents (Fig.~\ref{fig:priors}).

When deploying the IT\&E algorithm, the disparity between the quality of the prior map (that is, the skill of the \textit{other} agent) and the abilities of the agent has an impact on how quickly an ideal level is found. This implies that, for our approach to work well for human players, we would need agents that perform at a level comparable to them, which is feasible for many simpler games. In games where this is not feasible, human playtraces could be used to either build useful prior maps directly, or train humanlike agents, through e.g.\ imitation learning.

These results, and our methodology, leave room for discussion. The use of hand-crafted behavioral features in MAP-Elites comes at the risk of selecting the \textit{wrong ones} (that is, where maps that should exhibit similar difficulties lie further apart); in future work, the use of unsupervised feature learning (e.g. using Variational Autoencoders \cite{Thakkar2019, Summerville2018}) could be explored. Moreover, as in \cite{kaushik2020adaptive},  the most useful agent prior could be selected in an online fashion, i.e. during play. Furthermore, the proposed method can only select from the levels found by MAP-Elites. In future research, we plan to extend this to generating levels online, using the Gaussian Process to estimate the difficulty of the level.

Since our method can robustly model  what the agent finds difficult about the levels that are being served, we believe our method has the potential to perform Dynamic Difficulty Adjustment in domains in which understanding the player's abilities is useful (e.g. education \& rehabilitation games).

\section*{Acknowledgements}
We thank Julian Togelius for helpful discussions. SR thanks for support by Google (Google Faculty Research Award 2019). Funding for this research was provided by the Danish Ministry of Education and Science, Digital Pilot Hub and Skylab Digital.

\bibliographystyle{IEEEtran}
\bibliography{conference_101719.bib}

\end{document}